\documentclass[10pt,twocolumn,letterpaper]{article}

\usepackage[pagenumbers]{cvpr} 

\usepackage{graphicx}
\usepackage{amsmath}
\usepackage{amssymb}
\usepackage{booktabs}
\usepackage{comment}
\usepackage{enumitem}

\usepackage[pagebackref,breaklinks,colorlinks]{hyperref}

\usepackage[capitalize]{cleveref}
\crefname{section}{Sec.}{Secs.}
\Crefname{section}{Section}{Sections}
\Crefname{table}{Table}{Tables}
\crefname{table}{Tab.}{Tabs.}

\begin{document}

\title{Visually Grounded VQA by Lattice-based Retrieval}

\author{Daniel Reich, Felix Putze, Tanja Schultz\\
Cognitive Systems Lab., University of Bremen, Germany\\
 {\tt\small\{dreich,felix.putze,tanja.schultz\}@uni-bremen.de}}

\maketitle

\begin{abstract}

Visual Grounding (VG) in Visual Question Answering (VQA) systems describes how well a system manages to tie a question and its answer to relevant image regions. Systems with strong VG are considered intuitively interpretable and suggest an improved scene understanding. While VQA accuracy performances have seen impressive gains over the past few years, explicit improvements to VG performance and evaluation thereof have often taken a back seat on the road to overall accuracy improvements. A cause of this originates in the predominant choice of learning paradigm for VQA systems, which consists of training a discriminative classifier over a predetermined set of answer options. 

In this work, we break with the dominant VQA modeling paradigm of classification and investigate VQA from the standpoint of an information retrieval task. As such, the developed system directly ties VG into its core search procedure. Our system operates over a weighted, directed, acyclic graph, a.k.a. ``lattice'', which is derived from the scene graph of a given image in conjunction with region-referring expressions extracted from the question. 

We give a detailed analysis of our approach and discuss its distinctive properties and limitations. Our approach achieves the strongest VG performance among examined systems and exhibits exceptional generalization capabilities in a number of scenarios. 

\end{abstract}

\begin{figure}[ht]
  \centering
  \includegraphics[scale=0.22]{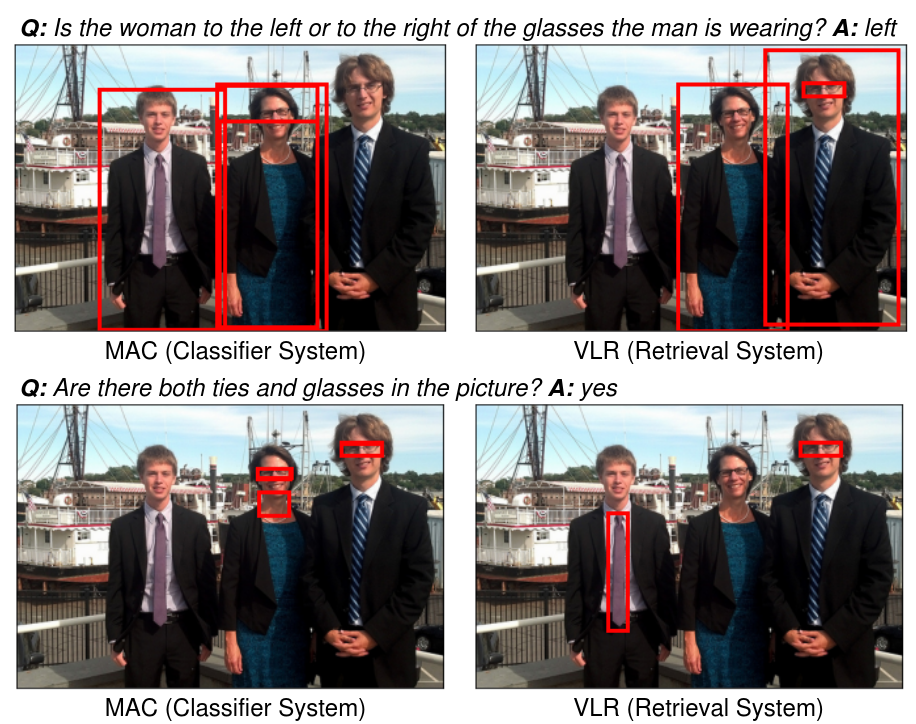}
  \caption{Illustration of Visual Grounding. Attended image regions of two VQA systems during inference. VLR's inference is correctly grounded while MAC focuses on parts that seem insufficient for producing the correct answer (which both systems do).}
  \label{fig:grounding_example}
\vspace{-4mm}

\end{figure}

\begin{figure*}[ht]
  \centering
  \includegraphics[scale=0.3]{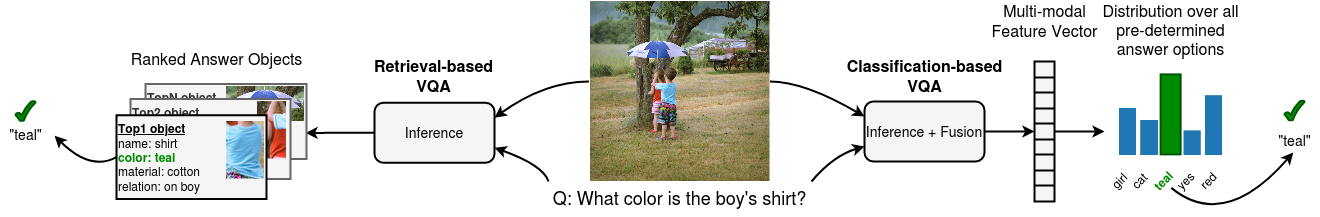}
  \caption{VQA system designs (simplified): The predominant classification design (r) and our lattice-based retrieval approach (l).}
  \label{fig:vqa_system_designs}
\vspace{-4mm}

\end{figure*}

\section{Introduction}
\label{sec:intro}
Visual Question Answering (VQA) \cite{vqa} deals with the multidisciplinary challenge of answering natural language questions about given images. As such, a VQA system requires proper handling of two types of inputs: 1) language, which encodes the query we seek to answer, and 2) images, which encode the search space for the query and act as knowledge base storing the answer. A third piece that completes this task is an inference engine that models the interaction between the two modalities and enables extraction of an answer. 
In this context, Visual Grounding (VG) can be described as a measure of how well the inference engine manages to tie region-referring expressions given in the question to relevant regions in the image and, consequently, produces an answer that is in fact based on these very regions. Systems with strong VG are considered intuitively interpretable (see Fig. \ref{fig:grounding_example}) and suggest heightened scene understanding. Improving VG in VQA makes models appear more trustworthy \cite{hint} and can improve accuracy (e.g. \cite{ying2022visfis}), while VG that is breaking apart can foreshadow a system's struggle in certain scenarios (see S\ref{sec:generalization}).

Some highly influential milestones in VQA system designs (e.g. \cite{bottomup_paper,SAN,neural_state_machine}), have been successful with designing their system's inference procedure closer to that of a human's inference process. While these designs might in theory implicitly enable a system to learn visually grounded inference, explicit efforts to improve VG performance and evaluation thereof have in practice taken a back seat on the road to overall accuracy improvements. Traditionally, VQA system designs have focused on creating a powerful discriminative classifier, often in the shape of an elaborate deep neural network, which is trained by minimizing an answer performance-related loss function like cross-entropy. 
Learning to produce correct answers via discriminative classification over a predefined answer set allows these models to identify complex patterns in the training data that help them select the correct answer. However, as image processing methods are still performing imperfectly in complex real-world scenarios and thus only produce sub-par image representations that VQA models have to use as (inaccurate) knowledge base to reason over, VQA systems may have to forego correct visual grounding for (seemingly random or unreasonable) patterns that will lead to correct answer selection (e.g. Fig. \ref{fig:grounding_example}). This line of reasoning suggests that building models with strong accuracy \textit{and} strong VG, or ``VQA for the right reasons'', is harder to accomplish than focusing on strong accuracy alone - especially for classifier-based systems. 

Accuracy has long been the primary performance metric driving development of novel VQA systems, with visual grounding being a ``nice-to-have'' property that is often overlooked in the presence of improved accuracy.
In this work, we turn this relationship on its head and treat VG as the primary metric to be improved, and investigate what new perspective a VG-driven VQA approach has to offer.
To this end, we address the challenge of ``VQA for the right reasons'' and design a VQA system that puts VG in the inference and answering process center stage. We break with the dominant VQA modeling paradigm of classification and investigate VQA from the standpoint of an information retrieval task (Fig. \ref{fig:vqa_system_designs}). 
Here, we frame the retrieval process of our system as a (scene) graph search problem and look to the field of Automatic Speech Recognition (ASR) for solutions, which has a long history of employing graph search algorithms for speech decoding. We adopt a concept that is integral to the ranking of recognition hypotheses in ASR: the word lattice.
Using symbolic features detected in the image (and represented in a scene graph), as well as inference instructions extracted from the question, we construct a weighted, directed, acyclic search graph, a.k.a. lattice. In this VQA-lattice, each path represents an alternative sequence of salient image regions, weighted by their recognition scores for object/attribute/relationship identities. The exact make-up of these scores depends on the region-referring expressions extracted from the question. Inspired by search procedures in ASR once more, we rank the paths through the VQA-lattice according to their likelihood using the Viterbi algorithm, and finally extract an answer via deterministic logic based on the 1-best path.

Approaching the VQA task in the manner described above manages to tie VG principles directly into our core search procedure for the answer, creating a highly visually grounded VQA system as evidenced by evaluations on the compositional-VQA-based GQA dataset \cite{gqa_dataset}. Moreover, following a retrieval-based paradigm enables our system to successfully handle generalization challenges that other classifier-based systems heavily struggle with.

\noindent\textbf{Contributions.} 
We summarize the contributions of this work as follows:
\begin{itemize}[noitemsep,nolistsep]
    \item A conceptually new VQA approach motivated by VG and based on information retrieval using lattices, named ``VQA-Lattice-based Retrieval'' (VLR).
\item We show that VLR achieves significantly stronger intrinsic VG than reference systems of various designs.

\item We examine VLR's distinctive strengths in various generalization and Out-of-Distribution scenarios, showing it is particularly well equipped for real-world deployment where such challenges are encountered.

\item VLR sets itself apart from other VQA systems by employing a quasi open-vocabulary answer production that is not restricted to a specially learned, pre-determined answer vocabulary.

\end{itemize}

\section{Related Work}
\label{sec:related}

\noindent\textbf{VQA System Designs.}
The modular system design that our system follows has been the foundation of numerous other VQA approaches. Various of these methods separate the reasoning process in a modular and explicit way by learning a program sequence that is subsequently executed on image features \cite{n2nmn,stacknmn,tbd,inferringprograms,pvr,zhao2021proto, chen2021meta}. Our approach differs from these in particular in the choice of image representation: we operate on scene graphs and use symbolic features instead of (spatial) visual features. 
Recent approaches increasingly employ scene graphs as image representation due to their usefulness in compositional VQA problems \cite{neural_state_machine,lcgn,liang-etal-2021-graghvqa,explainablesg,hypergraph,regat}. Our system shares similarities with the neuro-symbolic methods \cite{neural_state_machine}, \cite{pvr} and \cite{neuralsymbolicvqa} in particular, but is a fully modular system with inference path production and ranking capabilities (unlike \cite{neural_state_machine}, which is trained end-to-end and does not have an explicit mechanism to produce and rank inference paths) and operates on a probabilistic, symbolic scene graph representation as its knowledge base (unlike \cite{neuralsymbolicvqa}, which uses a discrete structural scene representation, and \cite{pvr}, which uses region-based visual features). All of these systems (with exception of \cite{neuralsymbolicvqa}, which queries a discrete, structured database for answers in an artificial, constricted scenario) have in common an answer production process that determines the output answer by means of a discriminative classifier, defined over a preset answer vocabulary. All mentioned systems are trained using an objective function that serves to directly improve answer performance. 

Finally, our system's retrieval-based design resembles DFOL, a method described in \cite{amizadeh2020neuro}, which proposes a formalism for (neuro-)symbolic reasoning in VQA. Our work distinguishes itself from DFOL in particular by 1) the introduced graph search concepts, 2) the more sophisticated Question Parser, and 3) our explicit conceptualization of VLR as a retrieval-based VQA system.

\noindent\textbf{Visual Grounding (VG) and Generalization in VQA.}
The problem of Q/A-biased learning in VQA systems leading to (or resulting from) a disregard of the visual modality has been discussed in work such as \cite{goyal2017makingv,vqacp}, leading to the introduction of datasets that aim to uncover such undesirable behavior by showing dramatic losses of accuracy performance in out-of-distribution (OOD) scenarios (for GQA: \cite{ying2022visfis, kervadec2021roses}). Various approaches have since been proposed to strengthen a model's (correctly grounded) reliance on the visual knowledge base (e.g. \cite{hint,Wu2019SelfCriticalRF,mutant}), thereby building on the motivation that improving VG in VQA will help improve a model's performance in generalization/OOD scenarios. In this context, however, \cite{shrestha-etal-2020-negative} showed that improvements to VG by mechanisms used in \cite{hint,Wu2019SelfCriticalRF} could not be proven to have been the primary cause of accuracy improvements on OOD tests. More recently, \cite{ying2022visfis} found that although their VG-related measurements on ID data were less predictive of OOD performance than ID accuracy, improving these VG-related metrics lead to accuracy improvements in both ID and (moreso) OOD tests. Accuracy boosts based on VG-related improvements were, however, tied to specific conditions, indicating that the accuracy-VG relationship is a complex one and still not fully understood.

Various systems focus on improving VG performance in the context of the GQA dataset, which we use as reference points in our evaluations: The module-based approach PVR \cite{pvr} improves VG by deriving explicit inference instructions (similar to our work's Question Parser) and including additional supervision for strengthening correct grounding during training of their classifier. Similarly, MMN \cite{chen2021meta} also uses a Question Parser and employs a scored alignment between predicted object importance and annotated relevant objects in each inference step as additional learning signal. MAC-Caps \cite{foundreason} proposes to use visual capsule modules \cite{capsule} for question-conditioned selection of detected image properties to improve VG performance. 

Lastly, we note that VG is also studied in non-VQA scenarios, e.g. in Referring Expressions research, where the goal is to locate textually described regions in the image, or generate them (see e.g. the popular RefCOCO datasets \cite{refcoco}). This localization task can be considered a sub-problem of VQA which differs in certain aspects: E.g., VQA requires inference to produce additional output after localization (=answers), questions often refer to non-existent regions (e.g. existence verification), and may not explicitly refer to any image region at all (weather, scene). Also, VG research in VQA is primarily concerned with a model's inherent ability to \textit{rely on} relevant regions during its inference and answering process.

\section{Dataset}
\label{sec:dataset}

Throughout this work, we use the ``balanced'' dataset from GQA \cite{gqa_dataset} to train and evaluate all of VLR's modules. GQA is a dataset created for compositional VQA on real-world images and contains a cleaner subset of images and scene graphs from Visual Genome \cite{krishnavisualgenome}. Questions were artificially generated based on these scene graphs. The dataset contains functional programs (describing multi-step inference operations) for all questions, which we use to train our Question Parser (see S\ref{sec:qp}). To evaluate VG performance, we use the included grounding information that connects questions to relevant regions in the image. 

Models in this work are trained on the ``balanced'' train subset from GQA, development sets were created by randomly splitting off data from this subset. All evaluations are based on the ``balanced'' val set.

\section{System description}
\label{sec:system}
\begin{figure*}[ht]
  \centering
  \includegraphics[scale=0.25]{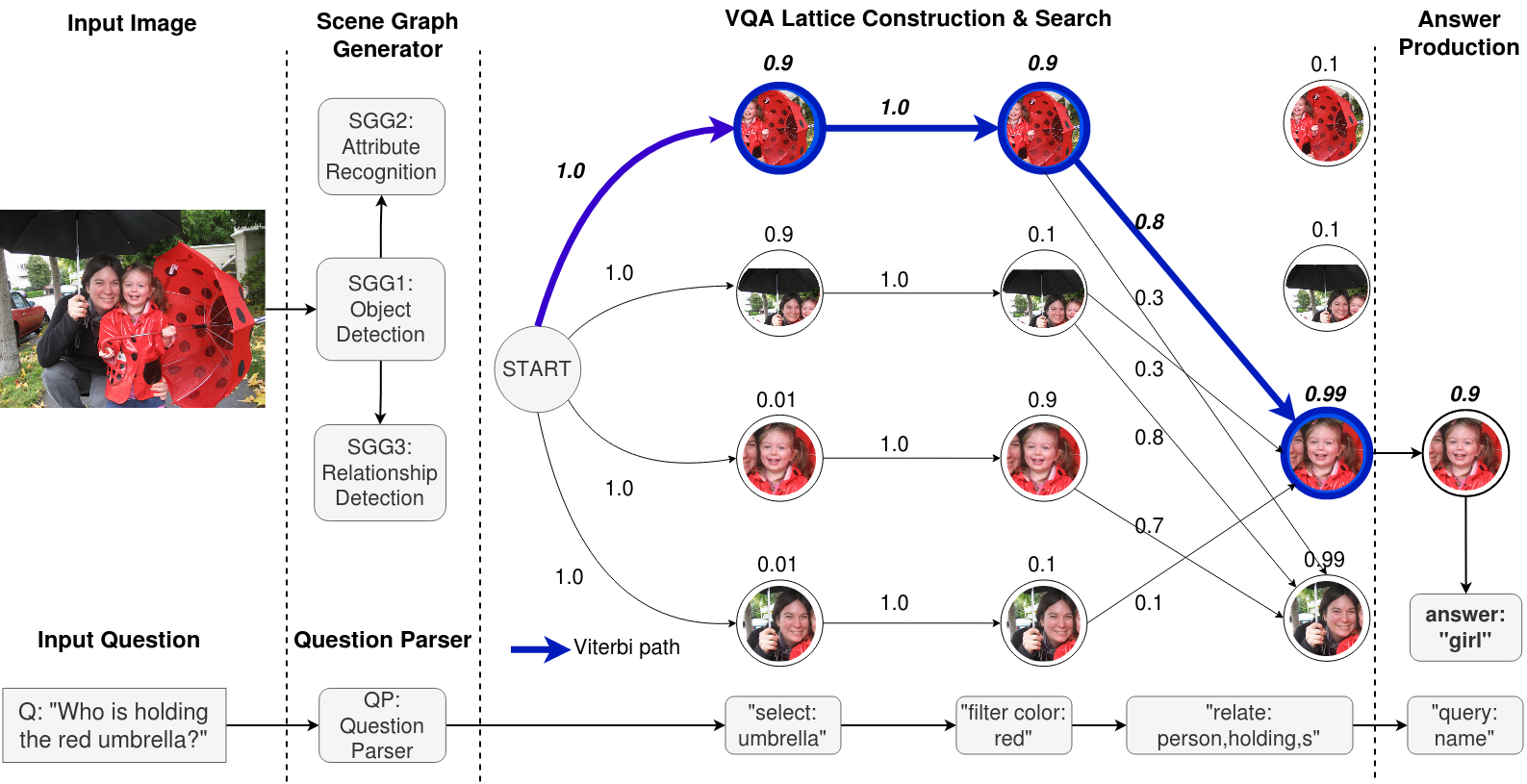}
  \caption{Overview of our system's components. VLR consists of a number of modules and steps (in bold), described in S\ref{sec:system}.}
\label{fig:overview}
\end{figure*}

VLR consists of three main modules (Question Parser, Scene Graph Generation, Rank \& Answer), which we describe in this section (overview shown in Fig. \ref{fig:overview}). 

\subsection{Question Parser}
\label{sec:qp}

The Question Parser (QP) parses the question into a sequence of operations (also called ``program'', cf. \cite{gqa_dataset,neuralsymbolicvqa}) that are used later in the construction of the VQA-lattice. In VLR, an ``operation'' is essentially a query to the scene graph to acquire specific recognition scores which are used in the construction of the VQA-lattice as node emissions or transition probabilities (see S\ref{sec:osp} for details). We use GQA-provided operation sequences (op-seqs) for training and evaluation of the QP.
An ``operation'' is a tuple given as (operation type, argument). Here, operation types take on values such as ``select'', ``filter'' or ``relate'', while arguments consist primarily of names for objects, attributes and relationships, as well as some functional symbols like logical-or and underscore. We determined 136 unique operation types in GQA, and a much higher number of arguments. These can be combined in various ways to form a myriad of unique operation tuples. 

\noindent\textbf{Model Description.}
The QP network is based on a pointer-generator architecture with ``coverage'', as presented in \cite{pointer_generator}. This network is, at its core, a seq2seq encoder-decoder model with attention mechanism and certain extensions that enable it to ``copy'' a token from the input to the output sequence. A ``coverage'' mechanism (\cite{coveragemechanism}) helps to significantly reduce token repetitions in the output sequence, which is a common issue in these networks. We choose this architecture for two main reasons: 1) To avoid an unnecessarily large output layer for generating all possible op-seq variants. As an operation tuple can refer to any object, attribute and relationship by name, the output vocabulary of a vanilla seq2seq model without the ``copy'' mechanism would have to be much larger to cover all those entity names, hence presenting a more challenging learning problem with higher data requirements, as well as stronger dependency on exhaustive program annotations. 2) The copy functionality of this network adds the capability to handle previously unseen entity names (e.g. new object names) reasonably well (see S\ref{sec:generalization}), whereas a vanilla seq2seq model (e.g. used in \cite{neuralsymbolicvqa,amizadeh2020neuro,zhao2021proto}) would not be able to output any word outside of its preset vocabulary.

\noindent\textbf{Model Specifications and Training.}
As model input, we use pre-trained GloVe word embeddings \cite{glove} to encode words in the question. 
The softmax output layer of the model consists of 162 classes, which includes a number of classes representing pointers to input question word positions.
We use a) regular expressions and b) GQA's annotated pointers (from question words to operation arguments) to determine whether or not a token in the target sequence should be a pointer to a certain question word at train time. Further details on the training process of the QP can be found in App. \ref{app_sec:qp}.

\subsection{Scene Graph Generation}
\label{sec:sgg}

The Scene Graph Generation (SGG) module is divided into three sub-modules to handle object detection, attribute recognition and relationship detection in an input image. We train a Faster R-CNN \cite{fasterrcnn} model for \textit{object detection and visual feature extraction}. This model's softmax output distributions (a 1702-dim vector for 1702 object classes per detected object) is used to populate the scene graph. An object's attributes (617 classes) as well as relationships among objects (310 classes) are similarly included in the scene graph. We defer to App. \ref{app_sec:scenegraph} for implementation details and in-depth performance evaluations.

\subsection{Rank \& Answer}
\label{sec:osp}
The Rank \& Answer module (R\&A) is responsible for determining the top ranked visually grounded inference paths and provides an answer to the given question. 
R\&A consists of three parts: 1) VQA-lattice construction, 2) Finding/ranking paths through the lattice, and 3) Producing the answer based on the best path.

\noindent\textbf{Lattice Construction.}
An automatic speech recognition (ASR) lattice is defined as a ``weighted, directed, acyclic graph in which each complete path represents an alternative hypothesis, weighted by its recognition score for a given utterance'' \cite{ljolje1999efficient}. Accordingly, we define a VQA-lattice as a weighted, directed, acyclic graph, where each complete path represents an alternative sequence of regions (objects) ending at the final answer region in an image, weighted by the image-based recognition scores from the scene graph as queried by the question. The source of recognition scores used in the VQA-lattice is determined by queries extracted from the question using the QP. These queries - or operation tuples - which consist of an ``operation'' and an ``argument'' (see also S\ref{sec:qp}), specify what the nodes in the scene graph should be queried about. For instance, the QP-extracted operation tuple (``select'', ``apple'') would query each node about its object recognition score for ``apple''. Similarly, (``filter color'', ``green'') would query each node about its attribute recognition score for ``green''. Queries about whole object categories like ``animal'' or ``furniture'' are handled by summing the recognition scores for all classes belonging to the category. Negations (e.g. ``not white'') are handled by subtracting the recognition score of the negated class from 1. Finally, operations involving relationships query the relationship-specific edges between any two nodes. 

\noindent\textbf{Rank.} 
With the lattice constructed, the search task can now be defined as finding the path through the lattice that maximizes the probability of the object/attribute/relationship detection models when applied to regions (objects) in the image according to the operations given by the QP. In contrast to ASR, where the goal is to find the maximum likelihood sequence of words given an audio signal, in VQA we want to find the maximum likelihood sequence of regions in an image, given both the image and a language-based description of the regions of interest (as given by the QP). We accomplish this, in accordance to ASR, by using the Viterbi algorithm, which is given as:
\begin{equation}\label{eq:eq1}
V_{0,r} = P(q_{0}|r) * \pi_{0}
\end{equation}
\begin{equation}\label{eq:eq2}
V_{t,r} = \underset{x \in I}{max} \left (P(q_{t}|r) * a_{x,r} * V_{t-1,x}  \right)
\end{equation}
where \textit{t} is the current inference step (=operation tuple), \textit{r} and \textit{x} are individual regions in the image, \textit{I} is the set of regions in a given image, $P(q_{t}|r)$ is the conditional probability of an image region matching a language-based region description from the QP (e.g. object or attribute identity) and \textit{a} is the transition probability from image region \textit{x} to \textit{r} (i.e. a relationship between them). Transition probabilities between image regions are based on QP output and provided by the relationship model (in SGG), if the current inference step describes a relationship between objects. If the current step does not involve relationships, $a_{x,r}$ is set to 1 if and only if \textit{x=r} and 0 otherwise. This forces the algorithm to stay with the same image region when processing subsequent attributional or positional descriptions of a queried object (e.g. a ``car'' that is ``red'').

The Viterbi path, i.e. the most likely sequence of image regions given the image and the question, is then retrieved from back pointers that remember the identity of the chosen region \textit{x} in Eq. \eqref{eq:eq2}:
\begin{equation}\label{eq:eq3}
x_{T} = \underset{x \in I}{V_{T,x}}
\end{equation}
\begin{equation}\label{eq:eq4}
x_{T-1} = BkPtr(x_{t}, t)
\end{equation}

with \textit{BkPtr} being a function that returns region \textit{x} used in Eq. \eqref{eq:eq2} for $t>1$ or \textit{r} if $t=1$. 

As is typical in information retrieval approaches, we produce a rank ordered list of the best path candidates after a search over the VQA lattice. For this, we use the parallel List Viterbi Decoding Algorithm (LVA) presented in \cite{seshadri1994list} to determine the n-best Viterbi paths in a given lattice.

\noindent\textbf{Answer.}
\label{sec:answerProd}
Our answer production module uses deterministic/programmed logic and contains no learnable parameters. To produce the answer to the question, our system depends on the final region(s) of the 1-best Viterbi path(s). 

GQA splits QA-pairs into five structural categories: query, choose, compare, logical and verify. Query type questions are also categorized as ``Open'' vocabulary questions, whereas all other types are categorized as ``Binary'' (with either ``yes''/``no'' answers, or the answer given as one of two options presented as part of the question). 
The extraction of the answer for ``query''-type questions, is a simple query on the scene graph for the class with the largest softmax score in the object’s class distribution (e.g. object or attribute name as defined by the scene graph). This query on the scene graph might be restricted to subsets of (object/attribute) classes depending on QP-extracted operations. E.g. the QP op-seq might have restricted the search to the category of ``furniture'' before querying the name of the final object, which allows a search space reduction to qualifying object names for the answer.
Note that some questions require the construction of two separate lattices and paths (e.g. ``logical-and'' type questions asking about existence of two separate objects). In these cases we produce one 1-best Viterbi path per lattice and then apply the final operation considering the ending nodes of both paths to answer the question.
Furthermore, ``verify''-type binary questions, which query object existence in the image, are answered by comparing the geometric average of all multiplied probabilities in the 1-best Viterbi path with a threshold value, which we determine on a dev set. 

In addition to the answer, VLR also returns the ranked Viterbi paths, providing the user with a highly transparent view at the inner workings of the QA-process.

\begin{table}[t]

\centering
\resizebox{\columnwidth}{!}{%
\begin{tabular}{lcccc}
\toprule
System & Binary & Open & Accuracy & Grounding \\
\midrule
N2NMN \cite{n2nmn} & n/a & n/a & n/a & 55.44 \\
UpDn \cite{bottomup_paper} & 74.60 & 47.30 & 60.51 & 94.42 \\
MAC \cite{mac} & 77.90 & 48.37 & 62.66 & 94.90 \\
PVR \cite{pvr} & 80.67 & 49.29 & 64.47 & 97.44\\
MMN \cite{chen2021meta} & \textbf{81.89} & \textbf{50.92} & \textbf{65.91} & 98.22 \\
DFOL \cite{amizadeh2020neuro} & 67.55 & 44.74 & 55.78 & 114.37\\
VLR & 69.94 & 46.17 & 57.67 & \textbf{126.61} \\
\bottomrule
\end{tabular}
} 
\caption{Accuracy and VG results, sorted by Grounding (see S\ref{sec:visgroundEval}, S\ref{sec:answerProd} about all metrics). 
N2NMN, PVR results on val set taken from \cite{pvr}, others were trained by us.}
\label{table:rawAcc}

\vspace{-4mm}
\end{table}
\section{Experiments}
\label{sec:experiments}

\begin{table*}[htbp]

\centering
\begin{tabular}{lccccccccccc}
\toprule
Line & System & QP & SG && Q & A & FA & Q+A+FA && Grounding & Acc \\
\midrule
1 & Det.Obj. & n/a & VLR && 91.45 & 92.34 & 91.66 & 91.49 && n/a & n/a \\
\midrule
 2 & UpDn \cite{bottomup_paper} & n/a & VLR && 23.72 & 39.32 & 28.78 & 29.90 && 94.42 & 60.51 \\
 3 & MAC \cite{mac} & n/a & VLR && 25.10 & 38.56 & 29.32 & 30.37 && 94.90 & 62.66\\
 4 & MMN \cite{chen2021meta} & MMN & VLR && 30.34 & 29.94 & 29.14 & 30.34 && 98.22 & \textbf{65.91}\\
 5 & DFOL \cite{neurosymbolic_reasoning} & GQA & VLR && 39.04 & 39.57 & 42.41 & 43.56 && 114.37 & 55.78\\
6 & VLR & VLR & VLR && \textbf{49.05} & \textbf{48.47} & \textbf{52.99} & \textbf{54.28} && \textbf{128.41} & 57.67 \\
\midrule
7 & VLR-2 & GQA & VLR && 57.47 & 31.90 & 55.11 & 61.23 && 132.00 & 60.16 \\
8 & VLR-7 & VLR & GQA    && 62.97 & 46.16 & 63.29 & 70.04 && 149.82 & 79.88 \\
9 & VLR-Oracle & GQA & GQA    && 70.93 & 50.94 & 70.75 & 78.46 && 162.73 & 91.78 \\
\bottomrule
\end{tabular}
\caption{VG results, discussed in S\ref{sec:visgroundEval}. VLR exhibits strong VG in all categories and metrics (L6). Line 1 shows per-question avg percentages of question-referred objects in the detected input scene graph. Best result in bold (only non-Oracle systems, i.e. L2-6).}
\label{table:visualGrounding_eval}

\vspace{-4mm}

\end{table*}

\subsection{General Evaluation}

We compare a number of VQA systems in T\ref{table:rawAcc}. With exception of N2NMN and PVR (results from \cite{pvr}), we trained all models with the same VLR-produced 1024-dim region-based visual features. MMN, PVR, DFOL and VLR use program generators (=question parser). DFOL runs on ground-truth programs from GQA, hence results for DFOL might be somewhat inflated (compare with VLR's ablations in App. \ref{app_sec:ablation} to gauge impact of using gt-programs under standard in-distribution (ID) testing conditions).

As shown in T\ref{table:rawAcc}, VLR's ID accuracy performance falls behind most of the reference systems, but surpasses all of them in VG. Given that there is no direct optimization of answer accuracy via minimization of a pertinent loss function involved in VLR, a lower ID accuracy is not surprising. VLR's focus on better VG, on the other hand, is clearly panning out and we investigate this and VLR's performance in various generalization scenarios and out-of-distribution (OOD) settings in more detail below.

\begin{table*}[!ht]

\centering
\begin{tabular}{lcccc}
\toprule
System & Objects & Ling. Variants & Answers & Low-Resource \\
\midrule
Train/Test & 763k/23k & 801k/20k & 862k/11k & 311k/132k \\
\midrule
UpDn \cite{bottomup_paper} & 35.75 (65.66; -45.6\%) & 58.20 (64.33; -9.5\%) & 0.0 (57.62; -100\%) & 49.72 (60.51; -17.8\%) \\
MAC \cite{mac} & 34.99 (64.91; -46.1\%) & \textbf{58.37} (64.05; -8.9\%)& 0.0 (57.96; -100\%) & 49.11 (62.66; -21.6\%) \\
MMN \cite{chen2021meta} & 41.71 (69.31; -39.8\%) & 57.27 (65.31; -12.3\%) & 0.0 (61.36; -100\%) & 51.24 (65.91; -22.3\%) \\
VLR & \textbf{56.40} (60.99; \textbf{-7.5\%}) & 52.19 (56.85;  \textbf{-8.2\%}) & \textbf{39.88} (51.63; \textbf{-22.8\%}) &  \textbf{51.58} (57.67; \textbf{-10.6\%}) \\
\bottomrule
\end{tabular} 
\caption{Generalization experiments, discussed in S\ref{sec:generalization}. Accuracy numbers in parenthesis for comparison when training in regular setting.} 
\label{table:refGeneralization}

\vspace{-4mm}

\end{table*}

\subsection{Visual Grounding Evaluation}
\label{sec:visgroundEval}

We first investigate VG performance in more detail and include measurements with an additional metric.

\noindent\textbf{Metrics and Categories.}
We use two metrics to measure VG performance: 1) the official ``Grounding'' metric from GQA (code at \cite{gqa_challenge}, description in \cite{gqa_dataset}), and 2) an IoU-based metric to measure how much attention a model puts on objects referenced in GQA's grounding annotations.
A third metric is used to compare VLR with MAC-Caps \cite{foundreason}, which is another model that focuses on improving VG (see App. \ref{app_sec:mac_caps} for results). We evaluate multiple metrics, because a) the GQA metric can create scores of $>100\%$ due to their algorithm summing attention scores for overlapping ground-truth regions, but we want to compare with existing work using the GQA metric like \cite{pvr,gqa_dataset}, and b) for additional verification of our results.

We calculate the IoU-based grounding scores as follows:
First, we check for each GQA-annotated inference object if any scene graph object has an IoU $>0.5$ with it. If yes, we add that scene graph object’s attention score to the final sum for that question.

GQA annotates inference objects for three categories: Objects referenced in the question (Q), in the short, one-word answer (A), and the full sentence answer (FA). We evaluate VG for each of these categories (as well as a combined one, ``Q+A+FA'') and calculate the IoU-based grounding scores as attention on inference objects in the category, averaged over all questions.

We base results on a single attention map for each compared model (UpDn, MAC, VLR, MMN, DFOL). UpDn produces a single attention map. For MAC, we use the map produced in the final reasoning step. Similarly, for VLR, we uniformly distribute 100\% attention among the final object(s) in the 1-best Viterbi path (as the answer is fully dependent on them). DFOL produces a relevance distribution over objects in each inference step. Here, we use the map produced for the final state (or average thereof for multiple final states). MMN uses a transformer-based architecture, employing multiple layers with multi-head attention. We take the avg of all attention maps involved in the inference process after their encoding layer (7 layers with 8 heads each for one inference step). We use the map representing the final inference step as MMN's map.

\noindent\textbf{Results discussion.}
Detailed results for evaluated systems are listed in T\ref{table:visualGrounding_eval}. IoU-based VG scores for categories Q, A, FA and Q+F+FA show that VLR features much stronger VG compared to other models in all categories, improving most scores by $>20\%$ rel. compared to the best performing reference system DFOL. Importantly, VLR is significantly more committed to relying on the \textit{final} answer region(s) of an inference path to produce its answer (T\ref{table:visualGrounding_eval}, column ``A''). VLR also outperforms MAC-Caps \cite{foundreason}, see App. \ref{app_sec:mac_caps}.
Results for various Oracle variants of VLR (T\ref{table:visualGrounding_eval}, L7-9) show that its grounding (measured over the entire inference path) improves further alongside accuracy, in particular with better scene graphs, while improvements in QP have less of an impact (e.g. going from VLR op-seq (T\ref{table:visualGrounding_eval},L8) to Oracle op-seq (T\ref{table:visualGrounding_eval},L9) shows only improvements of around 10\% relative in all grounding categories). This is consistent with trends we see for accuracy (see ablation study in App. \ref{app_sec:ablation}), and reaffirms the strong bond between perception modules and VG in VLR.

\subsection{Generalization \& Out-of-Distribution}
\label{sec:generalization_OOD}

VLR's compositionality and its retrieval-based design enable it to side-step some of the most prominent challenges that current classification-based VQA approaches struggle with, in particular certain content generalization and OOD scenarios. We investigate these scenarios in the following.

\subsubsection{Generalization Experiments}
\label{sec:generalization}
We perform experiments with re-partitioned train/test sets to investigate how systems handle four generalization settings that simulate challenges typically encountered in a practical real-world setting. Note that the underlying perception module (our SGG) remains unchanged for all systems in these experiments.
We give a quick overview of the settings here and defer to App. \ref{app_sec:generalization_ling} for more details. The data splits can be downloaded here\footnote{https://github.com/dreichCSL/GQA\_generalization\_splits}.

\noindent\textbf{Generalization to new object names.}
We test a model's ability to handle previously unseen object names in questions. Akin to \cite{neural_state_machine}, we remove all QA-samples from training that contain any object name from the food or animal category in the question. The test set then contains only these types of questions. 

\noindent\textbf{Generalization to linguistic variants.}
Similar to \cite{neural_state_machine}, we test generalization to variants of questions that are equivalent in terms of inference but are linguistically differently formulated (e.g. ``Do you see a car?'' vs ``Are there any cars in the image?''). Equivalent questions are re-partitioned such that we only train on questions of one linguistic variant and test on questions of the other variant.

\noindent\textbf{Generalization to new answer options.}
Unlike classifier-based approaches, VLR does not learn a pre-fixed answer vocabulary. This means that it can - in theory - produce an infinite amount of unique answers without retraining the system itself, while classifier-based systems are restricted to a pre-fixed set of answers. We illustrate this by removing all QA-samples with answers that are food or animal names from training and test on QA-samples with answers from only those categories.

\noindent\textbf{Low-resource training.}
To test a model's transferability, we simulate the massive input space of VQA under real-world conditions by causing a lack of exhaustive Question\&Image-Answer training tuples and disaligning train/test priors (similar to OOD conditions, but uncontrolled). We limit training samples per answer option to 1000 (test set remains unchanged).

\paragraph{\textbf{Results discussion.}}
Results for several systems are shown in T\ref{table:refGeneralization}. All models used in this section were trained with the same data. Note that DFOL, which uses ground-truth programs in its official code release, cannot be reasonably evaluated in these scenarios (no QP for retraining available).

VLR achieves the best accuracy in three experimental settings. Notably, VLR surpasses all shown models by large margins when generalizing to new objects and answers, which is a direct consequence of VLR's retrieval-based design and the QP's pointer-generator architecture that supports an open vocabulary (unlike e.g. MMN's parser).
For linguistic variants, most systems exhibit similar relative accuracy degradations (compared to a model trained on all data). In the Low-Resource setting, VLR suffers a much smaller relative drop than other systems. VLR performs well in this scenario, because it does not need to learn patterns between Q\&I-Answer pairs, it only needs to learn to accurately parse the question in isolation, which is a much less data-intensive task and not directly dependent on Q\&I-Answer coverage.

Measuring VG on the ``Object'' test, we observe significant drops of 14\%-28\% between regular and repartitioned settings for the classifier-based models, while VLR's Grounding only drops about 2\%. This provides some evidence that failing VG (relative to a model's VG standards) can foreshadow problems in answer performance. Note, that this is true for VLR in particular, where the relationship between accuracy and VG is especially strong (shown in VLR's ablation study in App. \ref{app_sec:ablation}).

\subsubsection{Out-of-Distribution Testing}
\label{sec:ood_experiments}

We now investigate performance of VLR in an OOD setting, introduced in  \cite{ying2022visfis} as GQA-101k. This data split is inspired by the popular VQA-CP set \cite{vqacp} and attempts to measure OOD performance in GQA. 
Evaluations in T\ref{table:ood_testing} show VLR's exceptional OOD performance on GQA-101k, with VLR's ID and OOD accuracies being virtually on par, while all other models exhibit a significant gap between their ID/OOD results. Measurements of Grounding are on avg almost equal for ID/OOD sets in intra-model comparisons, indicating that both sets share similar challenges in terms of VG. These measurements suggest that VG itself is not a major cause for the \textit{diverging} ID/OOD accuracy performances on this particular data split, but that the inference module in the answer production is likely failing on the OOD set due to over-reliance on the language component and the Q/A priors seen in training. VLR's performance shows that it can effectively nullify the ID/OOD accuracy gap by its indifference towards Q/A priors, further illustrating the advantages of its retrieval-based design.

\begin{table}[t]
\centering
\resizebox{\columnwidth}{!}{%
\begin{tabular}{lcc}
\toprule
System & GQA-101k \cite{ying2022visfis} & Grounding\\

\midrule
UpDn \cite{bottomup_paper} & ID: 55.21 OOD: 34.57 & ID: 81.20 OOD: 80.41 \\
MAC \cite{mac} & ID: 54.90 OOD: 33.08 & ID: 78.93 OOD: 79.10 \\
MMN \cite{chen2021meta} & ID: 54.84 OOD: 37.64 & ID: 75.08 OOD: 73.89 \\
VLR & ID: \textbf{55.55} OOD: \textbf{56.33} & ID: 124.34 OOD: 127.98 \\
\bottomrule

\end{tabular}
} 
\caption{Out-of-Distribution Testing.}
\label{table:ood_testing}
\vspace{-4mm}
\end{table}

\section{Limitations}
Performance-related limitations of VLR's modules have been analyzed in detail in VLR's ablation study (see App. \ref{app_sec:ablation}). 
In this section, we discuss design-related limitations of VLR regarding its application and extension to new use-cases.

\begin{enumerate}[noitemsep,nolistsep]
    \item Dependency on annotations: VLR learns from program annotations that map a question onto a functional notation defined for question types in GQA. Although we have shown VLR to have comparably lower requirements in terms of training data size, it still usually requires additional program annotations when retraining for new scenarios.
    \item Generalization to new question types: Related to (1).
    VLR can handle compositional questions used in GQA. Although an important subset of question types in the VQA-field fall into this category, VLR will struggle when faced with questions that significantly deviate from GQA in terms of program structure, thus requiring retraining (with additional annotations). Note that this point pertains to the \textit{program structure} of questions, which is different from VLR's ability to generalize well to new \textit{content} and - to a lesser degree - \textit{linguistic variants} in already learned question types.
    
    \item Handling sophisticated inference: As a retrieval-based system, VLR handles questions that can be answered directly based on retrieved paths involving objects/regions and their properties in an image. More involved inference, e.g. questions that require external world-knowledge to be answered, cannot be handled efficiently by VLR.
    
    \item Generalization to new answer types: Unlike classifier-based VQA systems, VLR can generalize to output answers it has never seen in training - but only if the answer \textit{type} is supported (e.g. returning the name of an object; confirming an object's existence). VLR uses deterministic logic in its answer production. If different types of answers need to be produced (e.g. counting objects), the answer production has to be manually expanded to add support.
    
\end{enumerate}

\noindent Note that for these reasons, VLR cannot be regularly trained \& evaluated on VQA datasets like \cite{vqa_challenge,goyal2017makingv,vqacp}.

\section{Conclusion}
We have introduced ``VQA-Lattice-based Retrieval'' (VLR), a modular, transparent system based on concepts from the field of Automatic Speech Recognition and Information Retrieval, that provides strong visual grounding for the task of compositional VQA.

In breaking with the traditional VQA learning paradigm of classification and designing a system from the standpoint of an information retrieval approach, we have shown that VLR achieves significantly better visual grounding in the inference process than various reference systems. We have also shown VLR's strong generalization capabilities in terms of handling new object names, linguistic variants, low-resource training and most notably its distinctive ability to produce answers never seen in training. VLR's exceptional performance in OOD scenarios further adds to its distinguishing strengths. Despite the discussed limitations of the approach and its shortcomings in ID accuracy compared to state-of-the-art models, VLR - as a retrieval-based system - offers some unique advantages that set it apart from other VQA systems. We can see VLR as a viable alternative to current classifier-based VQA systems in practical use-cases that require a) strong visual grounding (e.g. in order to handle follow-up queries; interaction with queried objects anywhere in the inference path; more explainable/predictable interaction), b) robust generalization performance in various scenarios, and c) an open answer vocabulary (e.g. for querying large and changing varieties of image contents). In terms of theoretical considerations, we hope that VLR and this study of its properties w.r.t. some of VQA's most prominent issues (VG, generalization, OOD scenarios) can inspire the formulation of new approaches in these areas.

{\small
\bibliographystyle{ieee_fullname}
\bibliography{cvpr_appendix_v02}
}
\clearpage

\appendix

\begin{center}
    \Large\textbf{Appendix}
\end{center}

\section{Scene Graph Generation}
\label{app_sec:scenegraph}
In this section, we give additional details for the Scene Graph Generation module described in the main paper, Section 4.2.
\subsection{SGG1: Object detection and visual feature extraction}
\label{app_sec:sgg1}
SGG1 handles object detection and provides region-based visual features that are used as basis for modeling attribute and relationship detection (as well as for training other reference VQA models). We use a Faster R-CNN \cite{fasterrcnn} model with ResNet101 \cite{resnet} backbone and an FPN \cite{fpn} for region proposals. We build the model using Facebook’s Detectron2 framework \cite{detectron2}. The ResNet101 backbone model was pre-trained on ImageNet \cite{imagenet_dataset}. 
We use the model's softmax output distributions (a 1702-dim vector for 1702 object classes per detected object) for up to 100 detected objects per image to populate the scene graph.

\paragraph{\textbf{Training details}}
We trained the model for GQA's 1702 object classes using 75k training images (images in GQA's train partition). Training lasted for 1m iterations with mini-batch of 4 images, using a multi-step learning rate starting at 0.005, reducing it by a factor of 10 at 700k and again at 900k iterations. No other parameters were changed in the official Detectron2 training recipe for this model architecture. Training took about 7 days on an RTX 2080 Ti.

\paragraph{\textbf{Output}}
We use the softmax output distribution (a 1702-dim vector for 1702 object classes) for each post-NMS detected object to populate our scene graph. Up to 100 objects per images are selected as follows: per-class NMS is applied at 0.7 IoU for objects that have any softmax object class probability of $>0.05$. 

We also use this model to extract 1024-dim object-based visual features, which we use in SGG2\&3 (attribute recognition, relationship detection models) and for training other VQA reference models which are used for performance comparisons.
We extract these visual features from a layer in the object classification head of the model which acts as input to the final fully-connected softmax-activated output layer. This is done for each surviving object (i.e. for the set of objects  determined with per-class NMS and top100 capping).

\paragraph{\textbf{Results}}
SGG1's object detection evaluation using metrics\footnote{\label{footnote:cocometric_foot}https://cocodataset.org/\#detection-eval} defined for the COCO dataset \cite{coco_dataset} are shown in Table \ref{app_table:sgg1}. We also include one more data point in Table \ref{app_table:sgg1} to share a very rough comparison to UpDn's \cite{bottomup_paper} object detection model that was used in many VQA-related works to extract high dimensional visual features. This model was trained on a heavily cleaned subset of Visual Genome \cite{krishnavisualgenome} (on which GQA is based) and uses 1600 object classes (results are unpublished, we print the ones listed in UpDn's official repository \cite{bottomup_repo}).

\begin{table}[t]

\resizebox{\columnwidth}{!}{%
\begin{tabular}{lcccc}
\toprule
System & mAP & mAP & mAP & mAP \\
& @$[.5:.95]$ & @0.5 & @0.75 & small/med/large  \\
\midrule
SGG1 & 5.54 & 9.45 & 5.76 & 3.32 / 6.13 / 9.92  \\
UpDn & n/a & 10.2 & n/a & n/a \\
\bottomrule
\end{tabular}
}  
\centering
\caption{ Object detection performances with Faster R-CNN models using COCO evalution metrics.}
\label{app_table:sgg1}
\end{table}

\paragraph{\textbf{Note on object categories}} Object categories take on a central role in the GQA dataset. Many questions refer to objects by their category instead of their object class identity (e.g. ``Is someone standing next to the red piece of \textit{furniture}?''). To determine category recognition scores for detected objects, we simply sum softmax scores of all object classes that belong to the requested category. Like for attribute recognition (see SGG2), we determine the categories, as well as the members (object class names) in each category, based on implicit declarations in GQA. Note that there is no explicit definition of the used categories and their members in GQA given as part of the official annotations, but this information can be inferred by appropriate processing of the Q/A annotation files. 

\begin{table}[t!]

\resizebox{\columnwidth}{!}{%
\begin{tabular}{lcccc}
\toprule
Attribute & Classes & Samples & Chance & Accuracy \\
Category & & train,dev,test & (=Prior) & \\
\midrule
activity & 15 & 11205,1244,1787 & 0.30 & 0.61 \\
age & 3 & 11610,1290,1754 & 0.46 & 0.75 \\
brightness & 2 & 11062,1229,1737 & 0.85 & 0.90 \\
cleanliness & 4 & 3887,431,596 & 0.58 & 0.70 \\
color & 26 & 456607,50734,71627 & 0.24 & 0.53 \\
company & 2 & 522,57,80 & 0.64 & 0.57 \\
depth & 2 & 420,46,64 & 0.62 & 0.69 \\
event & 3 & 143,15,20 & 0.60 & 0.70 \\
face & 4 & 3919,435,608 & 0.97 & 0.98 \\
fatness & 3 & 2078,230,327 & 0.57 & 0.53 \\
flavor & 4 & 1257,139,237 & 0.71 & 0.83 \\
gender & 2 & 887,98,154 & 0.68 & 0.76 \\
hardness & 2 & 562,62,90 & 0.54 & 0.80 \\
height & 2 & 16250,1805,2559 & 0.73 & 0.90 \\
length & 2 & 12920,1435,2041 & 0.67 & 0.84 \\
liquid & 5 & 608,67,94 & 0.53 & 0.67 \\
location & 2 & 579,64,71 & 0.82 & 0.86 \\
material & 41 & 67030,7447,10778 & 0.30 & 0.65 \\
opaqness & 2 & 7613,845,1076 & 0.99 & 1.00 \\
orientation & 2 & 393,43,68 & 0.53 & 0.65 \\
others & 427 & 141885,15765,22644 & 0.05 & 0.37 \\
pattern & 3 & 5085,564,893 & 0.79 & 0.79 \\
place & 5 & 647,71,101 & 0.31 & 0.75 \\
pose & 9 & 31410,3489,4774 & 0.38 & 0.64 \\
race & 2 & 656,72,96 & 0.62 & 0.62 \\
realism & 2 & 381,42,70 & 0.61 & 0.59 \\
room & 3 & 482,53,72 & 0.43 & 0.50 \\
shape & 5 & 12031,1336,1749 & 0.63 & 0.74 \\
size & 6 & 50809,5645,7779 & 0.54 & 0.72 \\
sport & 4 & 6705,745,955 & 0.61 & 0.97 \\
sportActivity & 8 & 8325,925,1233 & 0.28 & 0.78 \\
state & 6 & 3393,377,540 & 0.46 & 0.64 \\
texture & 2 & 19,2,9 & 0.78 & 0.56 \\
thickness & 2 & 2843,315,432 & 0.56 & 0.68 \\
tone & 2 & 10885,1209,1733 & 0.85 & 0.86 \\
type & 3 & 6442,715,917 & 0.64 & 0.97 \\
weather & 9 & 11477,1275,1691 & 0.63 & 0.76 \\
weight & 2 & 1980,219,307 & 0.85 & 0.84 \\
width & 2 & 1156,128,219 & 0.74 & 0.76 \\
\midrule
Avg (std) & - & - & 0.58 (0.14) & 0.73 (0.14) \\
Weighted avg (std) & - & - & 0.31 (0.21) & 0.59 (0.20) \\
\bottomrule
\end{tabular}
} 
\centering
\caption{Attribute recognition with softmax regression models, results sorted by category name. One model per category. See text (\ref{app_sec:sgg2}) for more details.}
\label{app_table:sgg2}
\end{table}

\subsection{SGG2: Attribute recognition}
\label{app_sec:sgg2}
We identified a total of 617 individual attribute names and 39 overarching attribute categories in GQA, including a category containing all un-assigned attributes (``others'' in Table \ref{app_table:sgg2}). We train a separate softmax regression model for each of these 39 attribute categories, using the visual features from SGG1 as input features. 
Classifier sizes range from a maximum of 427 classes (``other''
category) to a minimum of 2 classes (e.g. ``weight'', ``height''), with
most classifiers covering 3 or less classes. Categories and
category membership were determined based on their declaration in GQA.

\paragraph{\textbf{Training details}}
Input to each category model is a detected object's 1024-dim visual feature vector extracted with SGG1. For training and evaluation, the detected objects are labeled with attributes as follows: We first determine, if a SGG1-detected bounding box matches an (attribute-)annotated bounding box in GQA. If yes, we  assign any annotated attribute labels belonging to the ground-truth bounding box to that detected object. A detected bounding box has a match if it exceed an IoU of 0.75 with the ground-truth bounding box. The labeled samples are then used to train/evaluate every category model they have labels for.

We train all models using the Adam optimizer with a learning rate of 0.001 and apply L2 regularization to avoid overfitting. As loss function we use the common cross-entropy loss. Models are trained using early stopping with patience of 5 epochs.

\paragraph{\textbf{Output}} Attributes for detected objects in the scene graph are created as follows: we feed a detected object's 1024-dim visual feature vector into each of the 39 models and extract the softmax activation output distribution over each category's classes. The resulting 39 distributions then represent the complete attribute information of an object in the image's scene graph.

\paragraph{\textbf{Results}}
Results for all 39 category models, as well as sample sizes of train, dev and test sets are listed in Table \ref{app_table:sgg2}. 

\paragraph{\textbf{Note on attribute categories}} Like object categories, attribute categories are used extensively in GQA. In VLR, categories that are mentioned in the question (and/or later in the op-seq which is generated based on the question) can e.g. help narrow down answer options for a given question. E.g. a question like ``What \textit{color} is the chair?'' causes the Answer Production sub-module to select an answer from classes (attribute names) belonging to the ``color'' category. 
Like for object categories, attribute categories and category membership are determined based on implicit declarations in GQA. 
Note that a few attributes are part of multiple categories (e.g. the attribute ``little'' is a class in categories ``size'' and ``age''), which is why the total number of classes in all trained attribute models is slightly higher than the number of unique attribute names.

\subsection{SGG3: Visual relationship detection}
\label{app_sec:sgg3}
We identified 310 relationship names in GQA (e.g. ``wearing'', ``holding''), including 17+2 spatial relationships (e.g. ``behind''). Due to the overwhelming frequency of the two spatial relationship classes ``to the left$\vert$right of'' in the annotations (see counts in Table \ref{app_table:rel_recognition}), we place them in a separate category. Hence, we split all relationship names into three categories (spatial types, left-right, others).

Relationships are a directed connection between two objects, i.e. they do not have the commutative property (e.g. ``man wearing shirt'' is not the same as ``shirt wearing man''). We therefore frame relationship detection as a sequence classification problem. 

Using visual features and bounding box coordinates of detected objects from SGG1, we train an LSTM model with a softmax output layer for each of the three categories. We additionally train a binary classifier for each category that learns whether or not a given object pair has any relationship in the respective category. Input features here consist of GloVe word embedding vectors representing the detected object class names.

\begin{table}[t!]

\resizebox{\columnwidth}{!}{%

\begin{tabular}{lcccc}
\toprule
Relationship& Classes & Samples & Chance & Acc \\
Category & & train,dev,test & (=Prior) & \\ 
\midrule
Spatial & 17 & 69k, 8k, 11k & 0.27 & 0.55\\
Spatial (left$\vert$right) & 2 & 861k,96k,134k & 0.5 & 0.997 \\
Other & 297 & 55k, 6k, 9k & 0.40 & 0.77 \\
\bottomrule
\end{tabular}
}  
\centering
\setlength{\tabcolsep}{4pt}
\caption{Visual relationship detection. Results of relationship \textit{recognition} models per category. These models assume there is a relationship between two objects (given as an ordered pair) in the respective category and determine which one it is.}
\label{app_table:rel_recognition}
\end{table}

\paragraph{\textbf{Training details}}
For training and evaluation, detected object pairs from SGG1 are labeled with relationships as follows: We first determine, if a SGG1-detected bounding box matches an (relationship-)annotated bounding box in GQA. If yes, we assign any annotated relationship labels belonging to the ground-truth bounding box to that detected object, but \textit{only} if the target object in that relationship has also been matched by another detected object. A detected bounding box matches with a ground-truth bounding box if it exceeds an IoU of 0.75 with it. A labeled relationship (i.e. an ordered pair of objects) is used to train/evaluate the category model that contains the labeled relationship class.

For \textit{recognition} of relationships in the three categories, we train an LSTM, each with 512 units followed by a dropout layer (drop rate=0.3) and a softmax output layer. Input to the LSTMs are two object's 1028-dim vectors (1024-dim visual features, 4-dim bounding box coordinates, taken from SGG1). The vectors are ordered according to the directed relationship of the involved objects (i.e. subject\textrightarrow object).

For \textit{detection} of presence of a relationship for an ordered pair of objects (subject, object), we train LSTM models, which are architecturally similar to the recognition models, but use a sigmoid activation in the output layer instead of a softmax activation. We train one model per relationship category. In contrast to the recognition models, the input features consist of 100-dim GloVe word embedding vectors representing the SGG1-detected 1-best object name. The vectors are appended by the respective 4-dim bounding box coordinates. The intuition behind using word embeddings instead of visual features is that language-based semantics of relationships between objects might provide a richer basis to model prior probabilities of relationships between objects than the observed visual features.

Note that to train/evaluate the binary classifiers, we also need samples of the negative class, i.e. examples of object pairs that are not in the relationship category of the positive class. We select these negative samples from among object pairs that only have relationships in other categories but not in the category in question.

\begin{table}[t]

\begin{tabular}{lccccc}
\toprule
Relationship & Class & Samples & Prec & Rec & F1 \\
Category & & test & & & \\ 
\midrule
Spatial & pos & 10.5k & 0.40 & 0.85 & 0.55 \\
 & neg & 139k & 0.99 & 0.90 & 0.94 \\
\midrule
Spatial  & pos & 134k & 0.99 & 0.97 & 0.98 \\
(left$\vert$right) & neg & 15k & 0.81 & 0.94 & 0.87 \\
\midrule
Other & pos & 9k & 0.44 & 0.90 & 0.59 \\
 & neg & 141k & 0.99 & 0.93 & 0.96 \\

\bottomrule
\end{tabular}
\centering
\setlength{\tabcolsep}{4pt}

\caption{Visual relationship detection. Results of relationship \textit{detection} models per category. These binary models determine whether or not an ordered pair of objects has any relationship in that category. We list Precision, Recall and F1-score for a positively/negatively classified relationship detection in a category.}
\label{app_table:rel_detection}
\end{table}

\paragraph{\textbf{Output}} Similar to SGG1\&2, we use a model's softmax output to define a probability distribution over relationship classes between two objects. Note that a probability distribution is only generated with recognition models if the respective detection model outputs the positive class for a given ordered object pair. Otherwise, we set all values for that relationship category to zero in the scene graph representation.

\paragraph{\textbf{Results}} Results for recognition models are shown in Table \ref{app_table:rel_recognition}, results for detection models are shown in Table \ref{app_table:rel_detection}. 

\subsection{Scene Graph Representation}
\label{sec:sgr}
We represent the scene graph internally as a collection of matrices to improve computational efficiency in VLR as follows:

\paragraph{\textbf{Nodes}}
All objects in the scene graph are represented in a matrix of dimensions (\#obj\textunderscore in\textunderscore img, \#obj\textunderscore classes). Object attributes are represented in a matrix of dimensions (\#obj\textunderscore in\textunderscore img, \#attr\textunderscore categories, \#max\textunderscore attr\textunderscore class\textunderscore in\textunderscore cat). Each object class (1702 total) and attribute class (617 total across 39 categories) receives a softmax score from their respective classifier, which is then stored in these matrices. 

\paragraph{\textbf{Edges}}
All relationships are represented in a matrix of dimensions (\#obj\textunderscore in\textunderscore img, \#obj\textunderscore in\textunderscore img, \#rel\textunderscore classes). Each object in the image can be in a relationship (310 total across 3 categories) with another object (but not itself). Each relationship class receives a softmax score from the classifiers described in \ref{app_sec:sgg3}.

\begin{figure*}[t!]
  \centering
  \includegraphics[width=\linewidth]{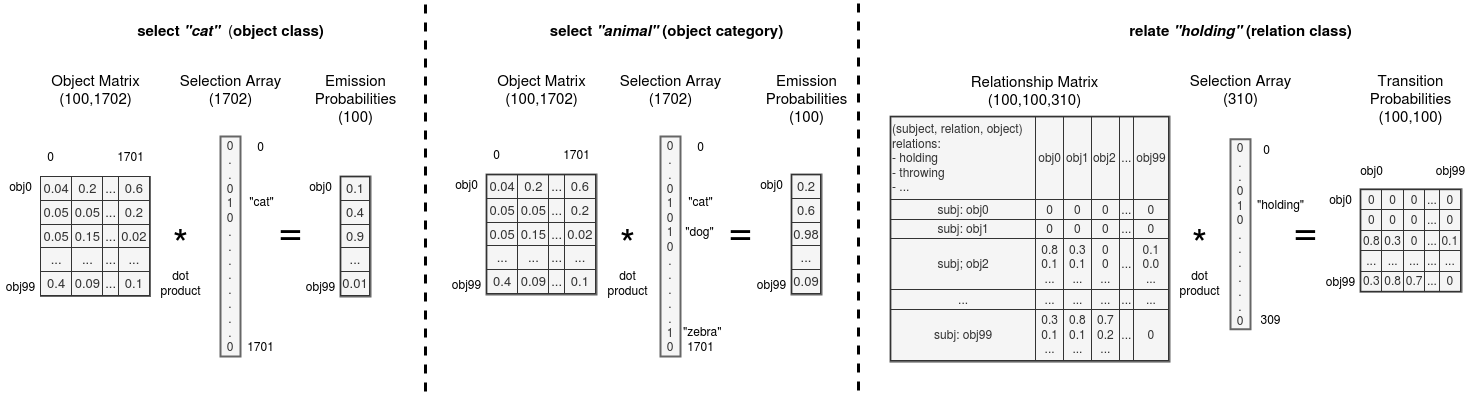}
  \caption{Illustration of elementary operations used in the construction of the VQA-lattice. Matrix operations are used to extract node emission and transition probabilities, which are subsequently used in the VQA-lattice. For more details see \ref{app_sec:lattice_constr}. Depicted numbers were randomly chosen.}
\label{app_fig:lattice_ops}
\end{figure*}

\section{System Description}

\subsection{Question Parser}
\label{app_sec:qp}
In this section, we give some additional details of the model setup and training process for our Question Parser described in the main paper in Section 4.1.
\subsubsection{Pre- and Post-processing}
\label{app_sec:qp_preprocessing}
For training the QP, we pre-process the operation sequences in a separate step, which in particular includes splitting each tuple (e.g. ``(relate; cat,next to,s)'') into multiple tokens. A post-processing step of the QP output is then required to revert the sequence into the original tuple format. Subsequent modules (namely R\&A in Section 4.3) process the op-seq in the original GQA tuple-based format. Note that there are instances where the pre-/post-processing steps introduce issues to the final QP output, which can cause problems during construction of the VQA-lattice. Therefore, we isolate and report on the impact of this step in our evaluation, as well (labeled ``VLR*'' in Table \ref{app_table:refVLRAblation}).

\subsubsection{Model specifications and Training}
We use 50-dim, pre-trained GloVe word embedding vectors to encode words in the question. As recurrent layers in the encoder/decoder, we use a GRU \cite{gru} with 128 units. Question length is limited to 20 tokens which matches 99.5\% of our training data. The total question counts before removal are 849k (train), 94k (dev) and 132k (test). Output sequences in the test set are 7.4 tokens long on average. The softmax output layer of the model consists of 162 classes (20 classes for pointers to question words, 1 class to signal empty/no operation, 93 (of 136) classes for operation terms that occurred $>100$ times in training, and the remainder for explicit modeling of the most frequently occurring (multi-) words in arguments, as well as non-word functional tokens like underscore, which never appear in the input question and thus cannot be copied from it). We use a) regular expressions and b) GQA’s annotated pointers (from question words to operation arguments) to determine whether or not a token in the target sequence should be a pointer to a certain question word in the input sequence at train time.

\subsubsection{Results}
The QP reaches 97.02\% element accuracy and 79.60\% full op-seq accuracy on the test set (GQA balanced val set: 132k questions with avg output sequence of 7.4 tokens per question op-seq). Note that the relatively high error for perfect parsing of op-seqs is not entirely indicative of its impact on inference path ranking and answer production in Section 4.3. For instance, changes in the order of operations (e.g. ``filter color: red'' followed by ``filter height: short'' and vice-versa) are counted as errors here, but should not matter for inference path ranking.

\subsection{Rank \& Answer}
In this section, we give some additional details about the inner workings of the lattice construction process described in the main paper in Section 4.3.
\subsubsection{Lattice Construction}
\label{app_sec:lattice_constr}
GQA uses 12 unique, fundamental operation types (e.g. ``filter'', ``select''; note: not to be confused with complex operation types like ``filter color'', etc), of which 9 (e.g. ``query'', ``and'') are used in VLR for determining the final operations in the answer production module that produces the answer to the question (see also main paper, Section 4.3, ``Answer''). 

VLR internally maps all operation types to (combinations of) two elementary query operations: ``select'' and ``relate''. These elementary operations are essentially queries to the Scene Graph Representation matrices and extract 1) node emission and 2) node transition probabilities, which are used in the VQA-lattice for a given question and image. An illustration of how these elementary operations work is shown in Fig. \ref{app_fig:lattice_ops} and described below.

An operation sequence (representing the parsed question) consists of two or more operation tuples. Each tuple consists of an operation type and an argument value (see also main paper, Section 4.1 and 4.3 ``Lattice Construction''). To run an elementary operation, we first use that argument value to create a ``Selection Array''. In Fig. \ref{app_fig:lattice_ops}, left, the argument is ``cat'' which results in the creation of a one-hot ``Selection Array'' that is 1 at the object class position for ``cat'' and 0 everywhere else. Taking the dot product of this one-hot vector and the object matrix then results in the node emission probabilities for all 100 detected objects in the image. The middle image in Fig. \ref{app_fig:lattice_ops} illustrates the same process for arguments that query object categories (which encompass multiple object classes). Queries about attributes are done in a similar fashion and also result in node emission probabilities for each detected objects. 

The second elementary operation (Fig. \ref{app_fig:lattice_ops}, right) extracts transition probabilities between nodes in the VQA-lattice in the shape of a hollow matrix (square matrix with diagonal entries all equal to zero), representing the fact that each node in the lattice can theoretically transition to any other node except itself.

Note that if node emissions are extracted by queries to the Attribute Matrix, we always insert the identity matrix to represent the transition probabilities of this step (Viterbi requires transition probabilities in each step), as no actual transition to other nodes is happening when attributes are queried.

Note also that VLR uses word-based symbolic representations for classes and categories. This means, in particular, that class names queried by operations need to match class names used in the SGG module, in order to get a correct match (or even a match at all). This strict requirement can be relaxed by introducing a normalization step before the matching procedure to align names in the query with the names used in the scene graph. We apply some normalization (lemmatization of plural/singular forms of object names), which treats most mismatches encountered with VLR in GQA. 
Although other mismatches are rare in GQA ($<1\%$ of questions), it might be helpful to include a normalization step in other scenarios.

\section{Experiments}



\subsection{VLR Ablation Study}
\label{app_sec:ablation}
The modular nature of VLR allows for an in-depth ablation-type evaluation.
We list a number of module combinations in T\ref{app_table:refVLRAblation}, which we discuss below.

\subsubsection{Preliminary Note on Scene Graph Variants}
\label{app_sec:note_scenegraph}
Rows in the ablation Table \ref{app_table:refVLRAblation} can be interpreted as representing the varying degrees of involvement of GQA annotations in the ablation (roughly: the higher the variant number, the larger the reliance on annotations). Because of the ablation-type study (Table \ref{app_table:refVLRAblation}) we have four scene graph variants, built to varying degrees of automatic detections vs annotations. For additional clarification, we list the four scene graph variants used in the ablation-type study in Table \ref{app_table:sgv}.

\begin{table} 
\resizebox{\columnwidth}{!}{%
\begin{tabular}{lcc}
\toprule
Scene Graph Variant & VLR-detected & GQA-annotated \\
\midrule
VLR, VLR-1, VLR-2 & obj, attr, rel & - \\
VLR-3, VLR-4 & obj, attr & rel \\
VLR-5, VLR-6 & obj & attr, rel \\
VLR-7, VLR-8, VLR-Oracle & - & obj, attr, rel \\

\bottomrule
\end{tabular}
}  
\centering
\setlength{\tabcolsep}{4pt}
\caption{Additional information on the scene graph variants used in the VLR ablation experiments in Table \ref{app_table:refVLRAblation}, see \ref{app_sec:note_scenegraph} for details.}
\label{app_table:sgv}
\end{table}

\begin{table*}[htbp!]

\centering
\begin{tabular}{lcccccc}

\toprule
System & QP & Scene Graph & Binary & Open & Grounding & Acc \\
\midrule
VLR & VLR & VLR (obj,attr,rel) & 69.94 & 46.17 & 126.61 & 57.67 \\
\midrule
VLR-1 & VLR* & VLR (obj,attr,rel) & 72.35 & 47.52 & 130.45 & 59.53 \\
VLR-2 & GQA & VLR (obj,attr,rel) & 73.58 & 47.59 & 132.00 & 60.16 \\
VLR-3 & VLR & VLR (obj,attr); GQA (rel) & 70.11 & 50.88 & 137.67 & 60.18 \\
VLR-4 & GQA & VLR (obj,attr); GQA (rel) & 75.18 & 53.40 & 141.19 & 63.93 \\
VLR-5 & VLR & VLR (obj); GQA (attr,rel) & 76.86 & 56.52 & 138.84 & 66.36 \\
VLR-6 & GQA & VLR (obj); GQA (attr,rel) & 82.99 & 59.65 & 142.23 & 70.95 \\
VLR-7 & VLR & GQA (obj,attr,rel) & 84.06 & 75.97 & 149.82 & 79.88 \\
VLR-8 & VLR* & GQA (obj,attr,rel) & 89.46 & 79.88 & 152.89 & 84.52 \\
VLR-Or & GQA & GQA (obj,attr,rel) & 96.47 & 87.38 & 162.73 & 91.78 \\
\bottomrule

\end{tabular}
\caption{Ablation study for VLR. Shows VLR's performance for various combinations of using annotated (=Oracle) and predicted scene graphs and operation sequences. ``GQA'' entries stand for Oracle inputs from GQA annotations. 
``VLR*'' is defined in \ref{app_sec:qp_preprocessing}. Here, we skip the learned QP but still go through pre- and post-processing (see \ref{app_sec:qp_preprocessing}) which introduces some errors. 
``VLR-Or'' represents VLR using full Oracle input (=GQA annotations), which acts as the upper bound of VLR.} 
\label{app_table:refVLRAblation}
\end{table*}

\subsubsection{Rank \& Answer (R\&A)}
The upper-bound performance of VLR when running in full Oracle mode (GQA-annotated scene graph and op-seq) is 91.78\% accuracy (T\ref{app_table:refVLRAblation}, VLR-Or). Problems occur e.g. due to 1) issues in annotations, or 2) when processing uncommon operation tuples that are not explicitly handled in R\&A's lattice construction procedures or answer production.

\subsubsection{Question Parser (QP)}
Using the GQA scene graph and QP-predicted op-seqs shows a steep drop in accuracy to 79.88\% or 13\% rel. (T\ref{app_table:refVLRAblation}, VLR-Or \textrightarrow VLR-7). Closer analysis reveals in particular problems w.r.t. inaccurate word pointers that cannot be correctly resolved, as well as general classification errors of the QP.
Comparing VLR op-seq with GQA op-seq for a fully VLR-detected scene graph (T\ref{app_table:refVLRAblation}, VLR-2 \textrightarrow VLR), this accuracy drop shrinks significantly (to 4\% rel.), suggesting a certain overlap between challenging visual scenes and more difficult questions.

\subsubsection{Object Detection (SGG1)}
Object detection is by far the most influential component in terms of accuracy impact. If an undetected object is queried in the question, VLR's ability to arrive at the correct answer is heavily impaired. 

VLR's SGG detects on average 91.49\% of all objects in GQA's annotated inference chain, and 92.34\% of objects needed to answer a question (detection determined for $IoU>0.5$). This means that a significant number of questions cannot be reasonably answered correctly on account of critical objects missing in the scene graph. All numbers are listed in the main paper (Table 2, L1). 

Aside from \textit{detection} issues, correct object classification (or \textit{recognition}) for a large number of objects like in GQA (1702 object classes) is clearly challenging (see also \ref{app_sec:sgg1} for object recognition results). 
To evaluate the impact on VLR's overall accuracy, we create the scene graph by using SGG's detected objects and retaining all annotated attributes and relationships for detected objects matching the annotations (``matching'': bounding box with highest IoU $>0.5$). As expected, this heavily impacts VLR's overall accuracy which falls from 91.78\% to 70.95\% (-23\% rel.) for Oracle op-seq (T\ref{app_table:refVLRAblation}, VLR-Or \textrightarrow VLR-6), and from 79.88\% to 66.36\% (-17\% rel.) for VLR-detected op-seq (T\ref{app_table:refVLRAblation}, VLR-7 \textrightarrow VLR-5).

\subsubsection{Attribute recognition (SGG2)}
Similar to object detection, attribute recognition is important for identifying objects referenced in the op-seq. We replace the GQA-annotated attributes in the partial Oracle scene graph from SGG1 (object detection) with outputs from the SGG2 module (attribute models). 
Although the impact of SGG2 on VLR's accuracy here is not as strong as for introducing detected objects from SGG1, we still see quite a large drop from 70.95\% to 63.93\% (Table \ref{app_table:refVLRAblation}, VLR-6 \textrightarrow VLR-4), and from 66.36\% to 60.18\% (Table \ref{app_table:refVLRAblation}, VLR-5 \textrightarrow VLR-3). 

\subsubsection{Relationship detection (SGG3)}
We integrate outputs from SGG3 (relationship models) into the scene graph to arrive at a fully VLR-generated scene graph. There is a smaller sized reduction in accuracy compared to the introduction of SGG2: from 63.93\% to 60.16\% (Table \ref{app_table:refVLRAblation}, VLR-4 \textrightarrow VLR-2) for Oracle op-seq, and 60.18\% to 57.67\% (Table \ref{app_table:refVLRAblation}, VLR-3 \textrightarrow VLR) to reach VLR's final overall accuracy without Oracle (annotation) involvement.

\subsection{Comparison with MAC-Caps}
\label{app_sec:mac_caps}
We include additional comparisions of VLR with MAC-Caps \cite{foundreason}, which attempts to improve VG in MAC \cite{mac} using visual capsule modules \cite{capsule}. We compare MAC-Caps with VLR using their F1-score-based metric (code at \cite{maccaps_github}). This metric measures Precision, Recall and F1-score of IoU@0.5-based matches between ground-truth objects and attended regions in the final step (see \cite{foundreason} for more details). 

Results are shown in T\ref{app_table:refVG_maccaps}. Note that MAC-Caps was designed and evaluated based on grid-features (i.e. not object-based features). Hence, both MAC and MAC-Caps numbers in this table are based on grid-features and were taken from \cite{foundreason} (and their Appendix). VLR uses object-based features, which gives it an additional advantage over MAC-Caps in object-centric VG evaluation, as illustrated by the large performance difference.

\begin{table}[!htbp]

\resizebox{\columnwidth}{!}{%

\begin{tabular}{lccccc}
\toprule
System & Prec & Rec & F1 & Ground & Acc\\
\midrule
MAC (from \cite{foundreason})& 1.97 & 2.28 & 2.11 & 41.68 & 57.09 \\
MAC-Caps \cite{foundreason} & 2.53 & 3.10 & 2.79 & 45.54 & 55.13 \\
VLR & \textbf{53.76} & \textbf{30.41} & \textbf{38.85} & \textbf{128.41} & \textbf{57.67} \\
\bottomrule

\end{tabular}
}  
\caption{VG comparison of MAC, MAC-Caps and VLR, using metrics (and results) from \cite{foundreason}, calculated for the final step of inference in the ``Q+A+FA'' category.} 
\label{app_table:refVG_maccaps}
\centering
\vspace{-4mm}
\end{table}

\subsection{Model Training}
\label{app_sec:model_training}
In this section we include details for training procedures of models we use in our evaluations. Generally, we use GQA's balanced train set to train all models and the balanced val set for evaluations. A small dev set (either some part of the train set or separately provided in case of experiments on GQA-101k \cite{ying2022visfis}) is used for model selection. Note that with exception of GQA-101k's test sets (which mix questions from balanced train and val sets), all images that are used in testing are unseen during training of our visual perception module (=SGG). All trained models use our same 1024-dim object-based visual features (for 100 objects/image max) as input.

\subsubsection{MMN}
MMN \cite{chen2021meta} consists of two main modules that are trained separately: A program parser and the actual inference model, which takes the predicted program from the parser as input. We mostly follow the settings in the official code-base but detail some aspects of our customization here. 

 For the program parser, we run training for 20 epochs (official setting: 10 epochs) and choose the best model (lowest loss on dev set). For the inference model, we run up to 15 epochs of bootstrapping (using the balanced train set) with Oracle programs and another up to 12 epochs of fine-tuning with parser-generated programs. We use early stopping of 1 epoch and select the model by best accuracy on the dev set (using Oracle programs in bootstrapping mode and predicted programs in fine-tuning mode).

Training on generalization/OOD splits is done accordingly. Notably, the program parser was always retrained on each new split (same as with VLR).

\subsubsection{DFOL}
DFOL \cite{neurosymbolic_reasoning} uses a vanilla seq2seq program parser, but neither code nor generated output for this is provided in the official code base. Thus, evaluations are run with ground-truth programs from GQA. DFOL is trained on a loss based on answer performance to learn weights in its visual encoding layers that produce an image representation similar to the one used by VLR, given high-dimensional visual features as input.

Training is done based on the official instructions for a complex 5-step curriculum training procedure. We train the first 4 curriculum steps with the entire ~14 million questions in the ``all'' training data partition, as specified in the instructions. As this is extremely resource intensive, we train for one epoch in each step. Finally, we run the 5th step with the ``balanced'' train data only (~1m questions) until training finishes by early stopping of 1 epoch.

Note that due to missing code for the program parser, DFOL cannot be reasonably evaluated in our generalization/OOD experiments, as the program parser's performance is key to a realistic evaluation for this model (similar to VLR).

\subsubsection{MAC}
MAC \cite{mac} is a monolithic VQA model based on a recurrent NN architecture which allows specification of the number of inference steps to take over the knowledge base. We follow the official training procedure guidelines given in the released code base and use 4-step inference. We train the model on GQA's balanced train set and use early stopping of 1 epoch based on accuracy on a dev set to select the best model.

Training on generalization/OOD splits is done accordingly.

\subsubsection{UpDn}
UpDn \cite{bottomup_paper} is a classic, straightforward attention-based model with a single attention step before merging vision and language modalities accordingly. We use the implementation shared by \cite{shrestha-etal-2020-negative}. Following the instructions there, we train UpDn for 40 epochs and select the best model based on accuracy on a dev set.

Training on generalization/OOD splits is done accordingly.


\section{Generalization Experiments}
\subsection{Generalization to linguistic variants}
\label{app_sec:generalization_ling}
For reproducibility, we describe in detail how we created the data partitions for this experiment (see main paper, Section 5.3.1 ``Generalization to linguistic variants'').
We tried to replicate the ``structural'' generalization experiment in \cite{neural_state_machine}, but found this to be infeasible with the short description given in the paper. Hence, we developed a process for re-partitioning the train/test data for this experiment.

The data splits can be downloaded here\footnote{https://github.com/dreichCSL/GQA\_generalization\_splits}.

\subsubsection{Determining question sets for each linguistic variant}
As starting point, we use the four explicitly mentioned linguistic variant pairs from \cite{neural_state_machine}, listed with examples in Table \ref{app_table:structuralExp}. We first determine all questions that belong to each linguistic variant for each of the four example pairs listed in Table \ref{app_table:structuralExp} as follows:
\begin{enumerate}
    \item Pair 1: Based on a question's (GQA-annotated) functional program, we determine all questions containing relationships names in passive form (``cover-ed'') vs present participle form (``cover-ing'').
    \item Pair 2: We determine all questions starting with ``Do/Does'' vs ``Is/Are''.
    \item Pair 3: We determine all questions containing an attribute category name (like ``material'', ``shape'', etc.) vs no such word.
    \item Pair 4: We determine all questions containing the word ``called'' vs ``name of''.
\end{enumerate}
At this point, we have two sets of questions for each of the four variant pairs. We do this procedure for the GQA balanced train set and val set separately.

\begin{table}[t]

\resizebox{\columnwidth}{!}{%

\begin{tabular}{cc}
\toprule
training & test\\
\midrule
What is the OBJ \textit{covered} by? & What is \textit{covering} the OBJ?  \\
\textit{Is} there a OBJ in the image? & \textit{Do} you see any OBJ in the photo?\\
What is the OBJ made of? & What \textit{material} makes up the OBJ? \\
What's the \textit{name of} the OBJ that is ATTR? & What is the ATTR OBJ \textit{called}? \\
\bottomrule
\end{tabular}
}  
\centering
\caption{Linguistic variant pairs used for re-partitioning the test/train set, akin to \cite{neural_state_machine}. For each pair (=row) we list examples of each of the two linguistic variants (=column). Questions belonging to a linguistic variant will be either in the new train or test partition.}
\label{app_table:structuralExp}
\end{table}

\subsubsection{Using program templates to determine equivalent inference}
The GQA-annotated functional programs can be generalized as templates by using placeholders in the program's arguments (e.g. for objects, attributes, relationships). For instance, ``(select: car), (verify color: red)'' can be generalized to ``(select: OBJ), (verify color: ATTR)''. We use these templates for generalized matching of inference chains between questions. To identify questions with equivalent inference programs, we now determine those templates that occur for questions in both variants of a variant pair. 
Note that we only select templates that occurred at least 100 times for each variant of a given variant pair. We found this threshold to be necessary because of incorrect program annotations in GQA that did not match the question. Without the threshold, this issue was compromising the experimental setup by causing the selection of large amounts of questions that did not actually have an equivalent inference match in the other variant group.

\subsubsection{Selecting questions for the new partitions}
After having determined the set of qualifying inference templates for each of the variant pairs, we can now create the new train and test partitions. Note that we do not move questions between GQA's balanced train set and val set, but create new partitions only by \textit{removing} questions from a set. This retains GQA's integrity of keeping only unseen images in the test partition (unseen in object detector training, i.e. our SGG).

We first select questions for each of the four linguistic variant pairs and then combine all selections to create the final train/test partitions.

\paragraph{Train set re-partitioning} For re-partitioning GQA's balanced train set, we select all questions with matching inference templates from the ``test''-labeled variant subsets in Table \ref{app_table:structuralExp} and remove them from the balanced train set. 
\paragraph{Test set re-partitioning} For re-partitioning GQA's balanced val set, we only select questions with matching inference templates from the ``test''-labeled variant subsets in Table \ref{app_table:structuralExp}.

\end{document}